\documentclass[runningheads]{llncs}
\usepackage{url}
\usepackage{xcolor}
\usepackage{graphicx}
\usepackage{amsmath}
\usepackage{amsfonts}
\usepackage{multirow}
\usepackage{booktabs}
\usepackage{xcolor}
\usepackage[export]{adjustbox}
\usepackage{algorithm}
\usepackage{algpseudocode}
\usepackage{tabularx}
\usepackage{ bbold }
\usepackage{hyperref}

\begin{document}
%
\title{A Generic Image Retrieval Method for Date Estimation of Historical Document Collections}
\titlerunning{Image Retrieval Method For Document Date Estimation}
%
\author{Adrià Molina\orcidID{0000-0003-0167-8756} \and Lluis Gomez\orcidID{0000-0003-1408-9803} \and Oriol Ramos Terrades\orcidID{0000-0002-3333-8812} \and Josep Lladós\orcidID{0000-0002-4533-4739}}
%
\authorrunning{A. Molina et al.}
%
\institute{Computer Vision Center and Computer Science Department, \\ Universitat Aut\`onoma de Barcelona, Catalunya\\
\email{\{amolina,lgomez,oriolrt,josep\}@cvc.uab.cat}}
\maketitle              
\begin{abstract}
    Date estimation of historical document images is a challenging problem, with several contributions in the literature that lack of the ability to generalize from one dataset to others. This paper presents a robust date estimation system based in a retrieval approach that generalizes well in front of heterogeneous collections. we use a ranking loss function named smooth-nDCG to train a Convolutional Neural Network that learns an ordination of documents for each problem. One of the main usages of the presented approach is as a tool for historical contextual retrieval. It means that scholars could perform comparative analysis of historical images from big datasets in terms of the period where they were produced. We provide experimental evaluation on different types of documents from real datasets of manuscript and newspaper images.



\keywords{Date Estimation \and Document Retrieval \and Image Retrieval \and Ranking Loss \and Smooth-nDCG.}
\end{abstract}

\section{Introduction}
    \label{sec:Introduction}
Universal access to historical archives and libraries has become a motivation to face innovative scientific and technological challenges. New services based in Document Analysis Systems are emerging for automatic indexing or extracting information from historical documents. In this context, the progress in Document Intelligence has brought the development of efficient methods for automatic tagging of such variety of documents. From a practical point of view, the association of metadata describing documents is extremely important in the early stages of archival processes, when archivists and librarians have to incorporate new collections (e.g. thousand new images of manuscripts, photographs, newspapers…). Metadata consists of annotated semantic tags that are used as keys in subsequent users’ searches. 
One of the most relevant information terms, in particular in historical documents, is the time stamp. Automatic dating has been particularly addressed in historical photos \cite{molina2021date,muller2017picture}. Visual features are highly important in dating of historical photographs. Texture and color features are good sources of information to accurately estimate when the image was taken, because photographic techniques have evolved. On another hand it is worth paying attention to the objects in the scene and their correspondence to some time periods (the clothes that people wear, their haircut styles, the overall environment, the tools and machinery, the natural landscape, etc.). But photographs are a particular type of historical assets, and the diversity is large. Date estimation is also interesting in mostly textual documents. A particular case is date estimation in manuscripts where writing style features and script recognition are crucial to classify a document within a time period or a geographic area (Hebrew, Medieval…) \cite{hamid2019deep,Dhali2020}. Hybrid methods combining multimodal features like appearance-based features and textual labels not only from the image itself but from its context when it appears in a document like a newspaper or a web page \cite{martin2014dating} have also demonstrated interesting performance. An interesting observation in the date is an ordinal feature, i.e. in some cases the date of a document can be better estimated by comparison to other ones. Ordinal classification techniques have been proposed so the task is to predict a ranked list of documents \cite{martin2014dating,molina2021date}.

The existing date estimation methods have two main drawbacks. First, they are highly dependent of having a priori and contextual knowledge, and second they are conditioned by the data target. Technically, it means that a lot of training data, labeled by experts is required. On another hand, from a systemic perspective, i.e. date estimation as a service to scholars and archivists, there is a need of genericity and adaptability. In a practical situation, an archivist or a librarian receives diverse collections to classify with bunches of images. They require tools able to deal with heterogeneous data collections. Additionally, the response the archivist or social scientist is looking for, is not specific or affordable for any estimator. This is where retrieval plays a key role: Some professional may be looking for the date of an image, where an estimator makes its work properly; but other may be looking for cues, patterns or historical keywords (such as a word itself, a cool hat or a particular historical character) that belong to a certain period. Estimators fails at providing this service. Using ranking functions for this task, should pave the way to generate embedding spaces where communities are constituted by its contextual information such as the date instead of the visual and textual information.

For the discussed above reasons, in this paper we present a date estimation approach that follows a ranking learning paradigm in a retrieval scenario. We propose an incremental evolution of our previous work on date estimation for photographs collections \cite{molina2021date}. In this work we proposed a model based in the Normalized Discounted Cumulative Gain (nDCG) ranking metric. In particular, the learning objective uses the {\em smooth-nDCG} ranking loss function. In the current work, we extend the functionality of the method, making it able to generalize to other types of documents. We will show its performance to estimate the date of two categories of documents: handwritten and historical newspapers joining the previously explored scanned photographs system in \cite{molina2021date}. As additional feature of the proposed system, we integrate the human in the loop, so the user herself/himself if able to integrate her/his own feedback for fine tuning the model.

The overall contributions of our work can be summarized as follows:

\begin{itemize}
\item A flexible date estimation method based in a ranking loss function which is able to estimate the date of a document image within an ordering of the whole collection. The method is adaptable to different types of documents.
\item The capability of generating embedding spaces according to contextual information such as dates instead of using textures, visual cues or textual ones.

\item An application that is able to deal with document collections of different types. In particular we will show the performance on historical newspapers and manuscript documents.
\item The inclusion of the user feedback in the process; making the application able to focus on those categories the user is currently interested on.
\item The evaluation on different real datasets, some of them provided by the Catalan National Archives Department.
\end{itemize}

The rest of the paper is organized as follows: in section \ref{sec:sota} we review the state of the art in date estimation and we state the contribution of this paper with respect previous work. The data used for the experimental validation and the details of the peculiar loss function are exposed in sections \ref{sec:data}, \ref{sec:learning_obj} respectively. In section \ref{sec:arch} the retrieval system is described. In section \ref{sec:experiments} we present the resulting application architecture and the quantitative results in terms of regression. We conclude with section \ref{sec:conclusions} where we draw the conclusions.

    

    

\section{Related Work}
    \label{sec:sota}
Before the widespread use of deep learning for document analysis tasks, automatic dating of historical manuscripts and printed documents was primarily based on hand-crafted features, carefully designed to capture certain characteristics of the handwriting style that are useful for identifying a particular historical period. Some examples of such hand-crafted features are Fraglet and Hinge features~\cite{he2014towards,bulacu2007text}, Quill features~\cite{brink2012writer}, textural measures~\cite{hamid2018historical}, histogram of strokes orientations~\cite{he2016Multiple}, and polar stroke descriptors~\cite{he2015polar,he2016historical} among others.

In recent years, thanks to the latest developments in deep learning, there has been a greater tendency to use models in which features are learned directly from training data to perform various document analysis and recognition tasks~\cite{lombardi2020deep}. In that regard, Li~\textit{et al.}~\cite{li2015date} proposed a custom Convolutional Neural Network (CNN) consisting of two convolutional layers and two pooling layers for the task of publication date estimation for printed historical documents. They evaluated this model in two different tasks: date regression and century classification (with four classes that span for 100 years each).

Wahlberg~\textit{et al.}~\cite{wahlberg2016historical} used a more modern CNN based on the GoogleNet architecture~\cite{43022} either for directly estimating the date of historical manuscripts or as a feature extractor in combination with other regression techniques. In particular, they explore the combination of CNN extracted features with a Gaussian Processes Regressor and a Support Vector Regressor. In this case the model is pre-trained on the Imagenet dataset~\cite{deng2009imagenet} and fine-tuned with a single output neuron for date regression. They extract 20 random crops of $256\times256$ from each manuscript, and at inference time they take the median of the 20 date estimates. A similar approach, using a pre-trained deep CNN model, was explored by Hamid~\textit{et al.}~\cite{hamid2019deep} and Studler~\textit{et al.}~\cite{studer2019comprehensive} but they evaluated several state of the art CNN architectures including VGG19~\cite{simonyan2014very}, GoogleNet~\cite{43022}, ResNet~\cite{he2016deep}, Inceptionv3~\cite{szegedy2016rethinking}, InceptionResnetv2~\cite{szegedy2017inception}, and DenseNet~\cite{huang2017densely}. Overall, their experiments demonstrated that ImageNet pre-training improves the performance of all the architectures, and that the patch-based ensemble (taking several crop patches of a manuscript) provides better results than using the whole document for both regression and classification tasks.


All the models mentioned above treat the problem of document dating either as a classification or a regression task. In this paper we propose to use different approach and recast the dating problem as a retrieval one. As shown in our previous work \cite{molina2021date,riba2021learning}, it is possible to train a neural network to learn document representations in an embedding that preserves a desired metric as input; in this case, the proportion between different dates (years) is preserved in the output space. This new approach allows us to include in the results a retrieval document set ranked by date similarity, which, again, is contextual information rather than strictly visual; despite context is deducted by visual clues, we're not optimizing for looking certain clues, but a general embedding that better fits the representation of a certain period with respect other ones. This is specially desirable in the proposed application of a support tool for professional archivist and historical researchers (Section~\ref{sec:experiments}) since it doesn't just return a categorical result but a set of similar labeled data that may help to get the proper conclusions by comparing the query and output data.

\section{Datasets}
    \label{sec:data}
Several document dating databases exist for both historical manuscripts~\cite{he_sheng_2016_1194357,cloppet2017icdar2017,adam2018kertas} and printed documents~\cite{vincent2007google}. In our experiments, we used two datasets: the Medieval Paleographic Scale (MPS) dataset and a custom historical newspaper dataset. 

\subsubsection*{Manuscript Date Estimation}\hspace*{\fill} \\

\begin{figure}
    \centering
    \includegraphics[height=4cm, width = 9cm]{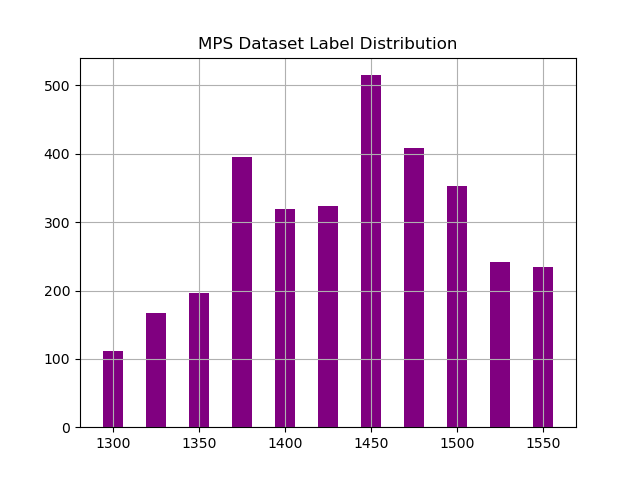}
    \caption{MPS Dataset histogram of images per label.}
    \label{fig:mps_histogram}
\end{figure}

One of the problems we tested our method on, is the date estimation of Medieval Paleographic Scale manuscripts, MPS Dataset, presented by He Sheng \textit{et al.}  \cite{he_sheng_2016_1194357}. The dataset presents several manuscripts with Dutch and Flemish manuscripts from medieval mid period, years 1300 to 1550, divided in steps of 25 years. The dataset is not uniformly distributed with respect the years Figure \ref{fig:mps_histogram}, which means that we lack of manuscripts from early and latest years.

\begin{figure}
    \centering
\begin{tabular}{c c c}
    \includegraphics[height=0.24\linewidth ]{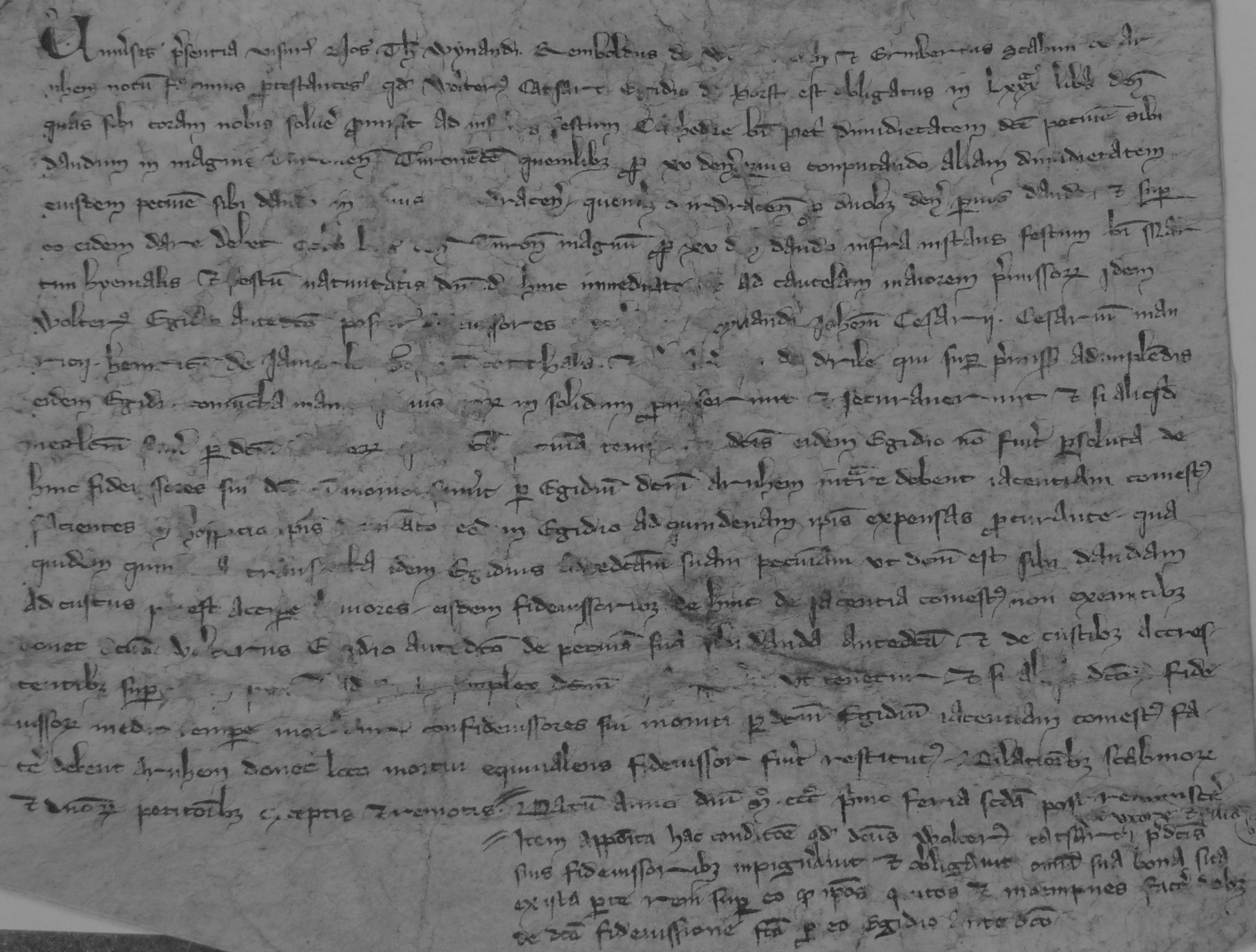} & 
    \includegraphics[height=0.24\linewidth ]{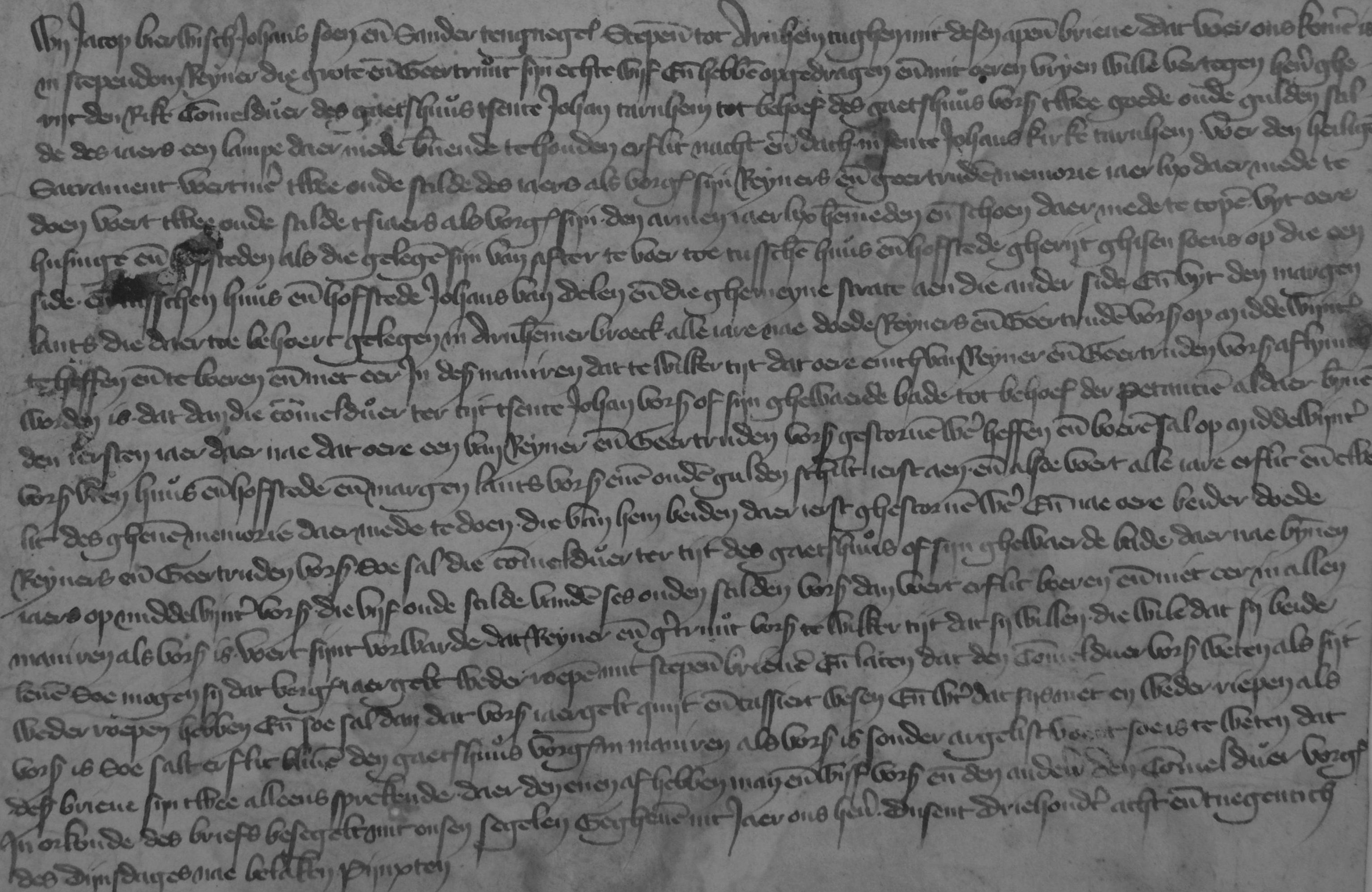} & 
    \includegraphics[height=0.24\linewidth ]{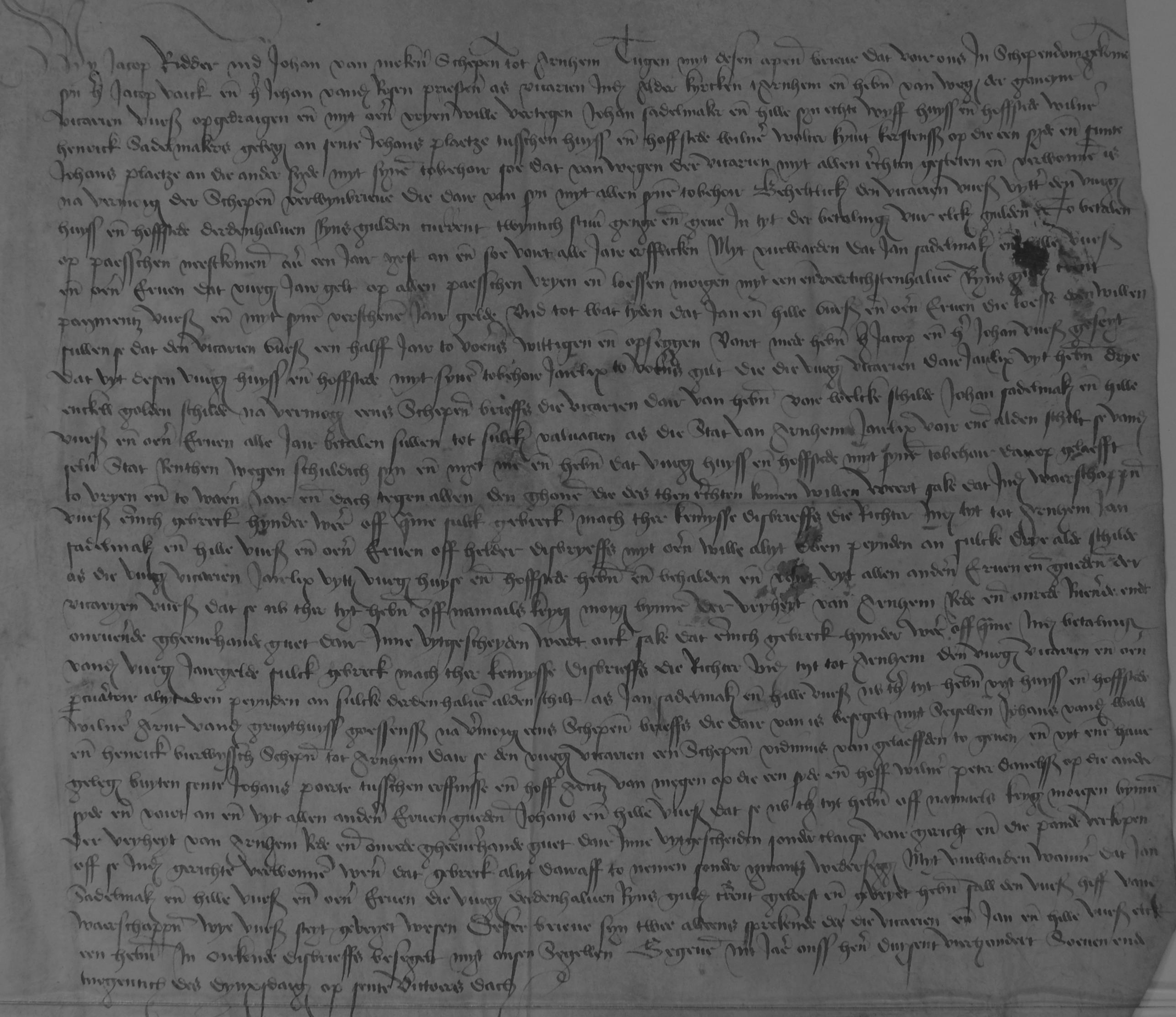}
\end{tabular}
    \caption{Sample of manuscripts extracted from the Medieval Paleographic Scale (MPS) dataset \cite{he_sheng_2016_1194357}. Images belong to years 1300, 1400 and 1500 respectively.}
    \label{fig:char}
\end{figure}

As introduced by He \textit{et al.}; we can observe a flagrant evolution in several characters in the dataset; leading to great results at character-level \cite{HE2016159} \cite{he2016Multiple}.
Since the dataset provides directly the whole manuscript, we use this as input data \ref{fig:char} for a more general view. 

\subsubsection*{Newspaper Date Estimation}\hspace*{\fill} \\

\begin{figure}
    \centering
    \includegraphics[width=8cm, height = 4cm]{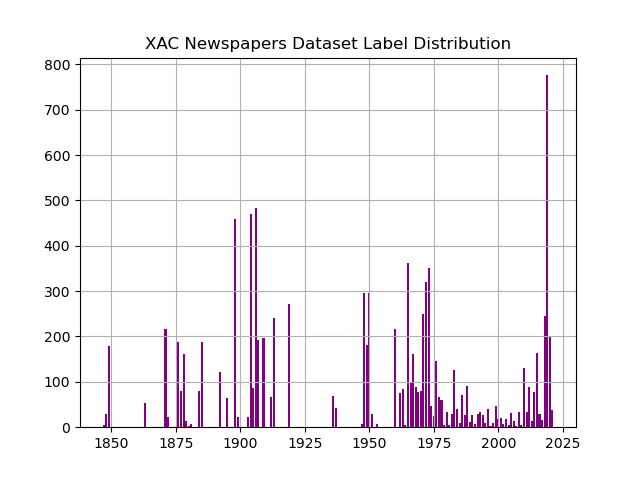}
    \caption{XAC Newspaper Dataset: histogram of images per label.}
    \label{fig:xac_histogram}
\end{figure}

\begin{figure*}
\begin{multicols}{3}
    \includegraphics[width=0.84\linewidth]{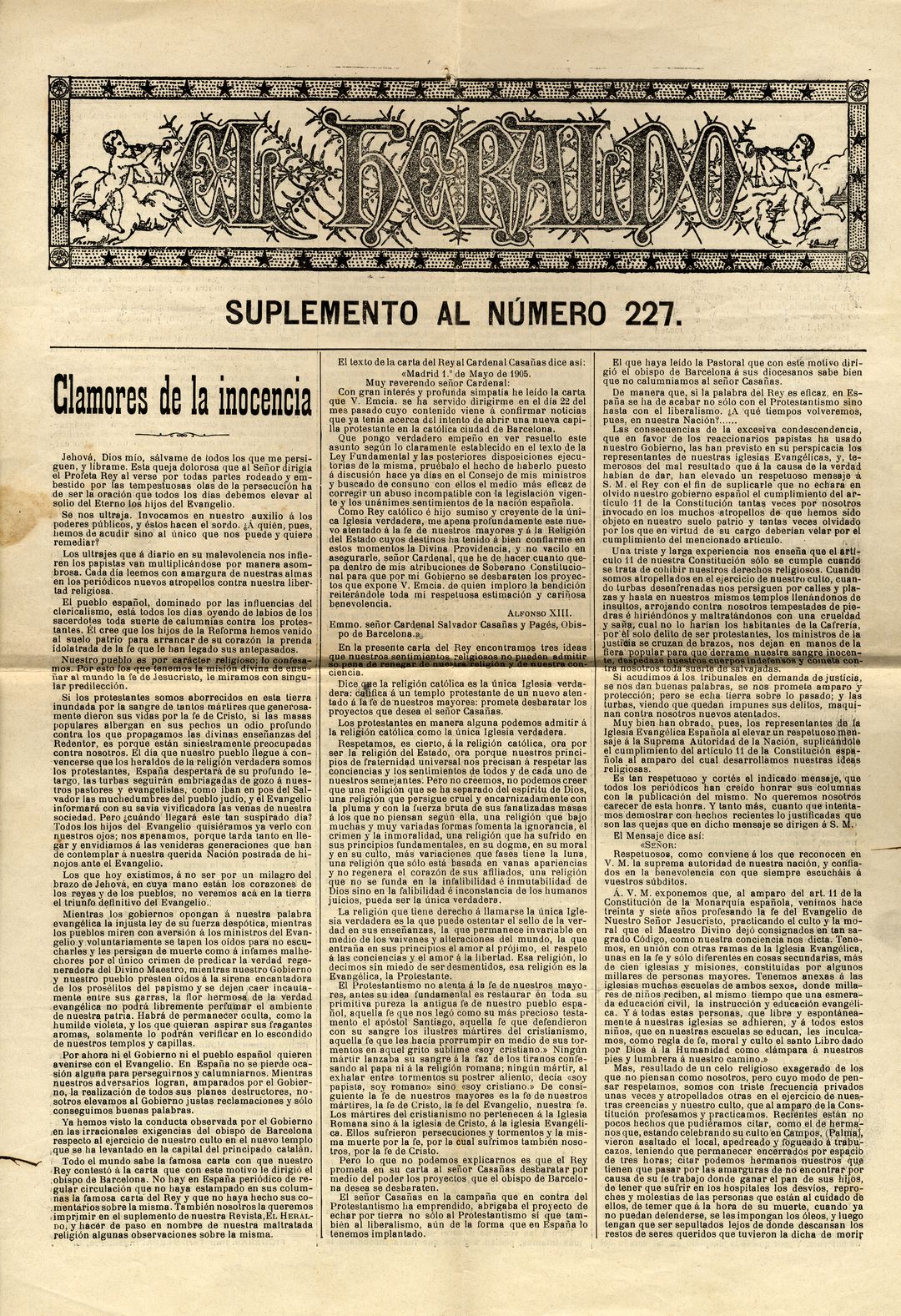}\par 
    \includegraphics[width=0.84\linewidth]{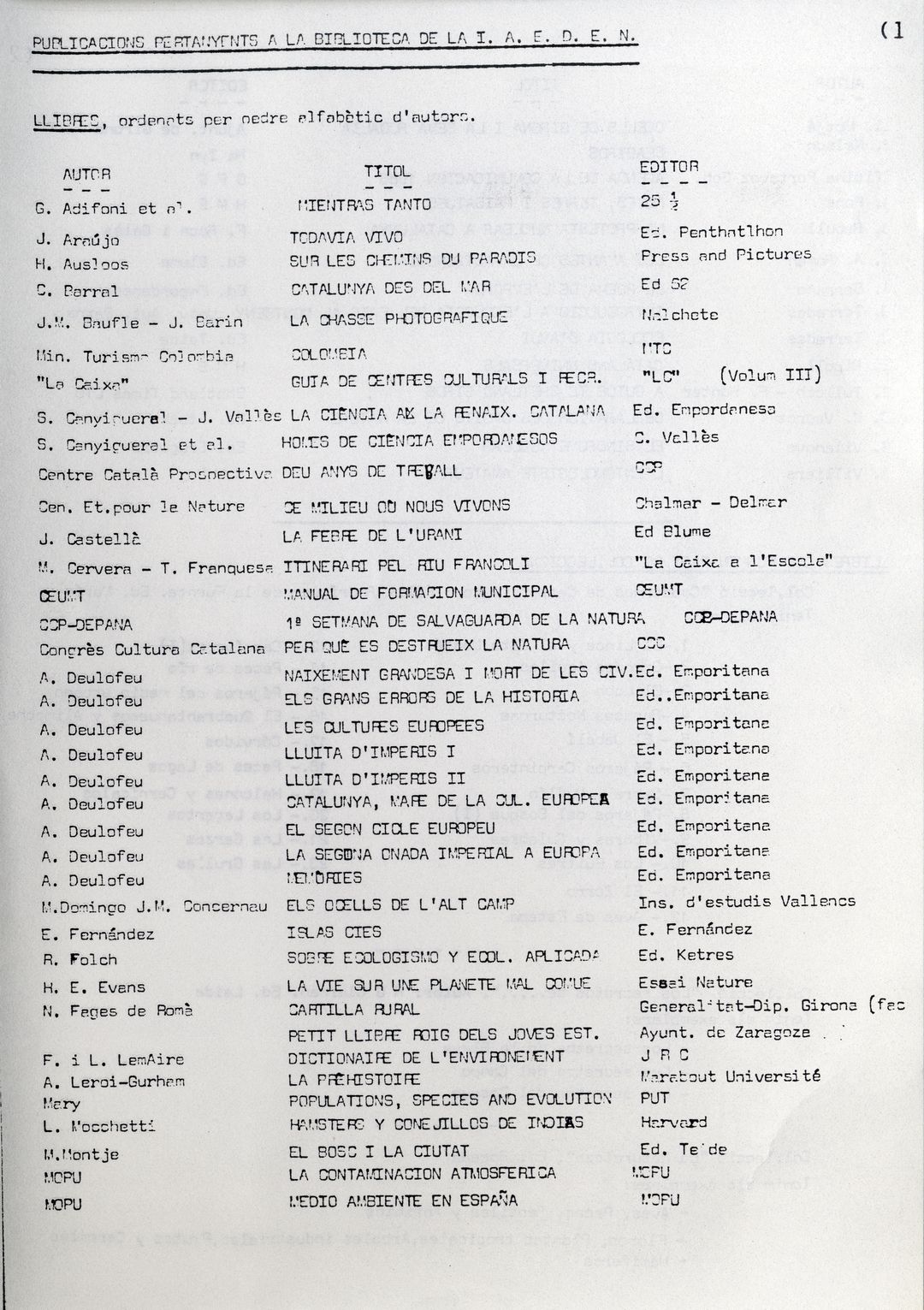}\par
    \includegraphics[width=\linewidth]{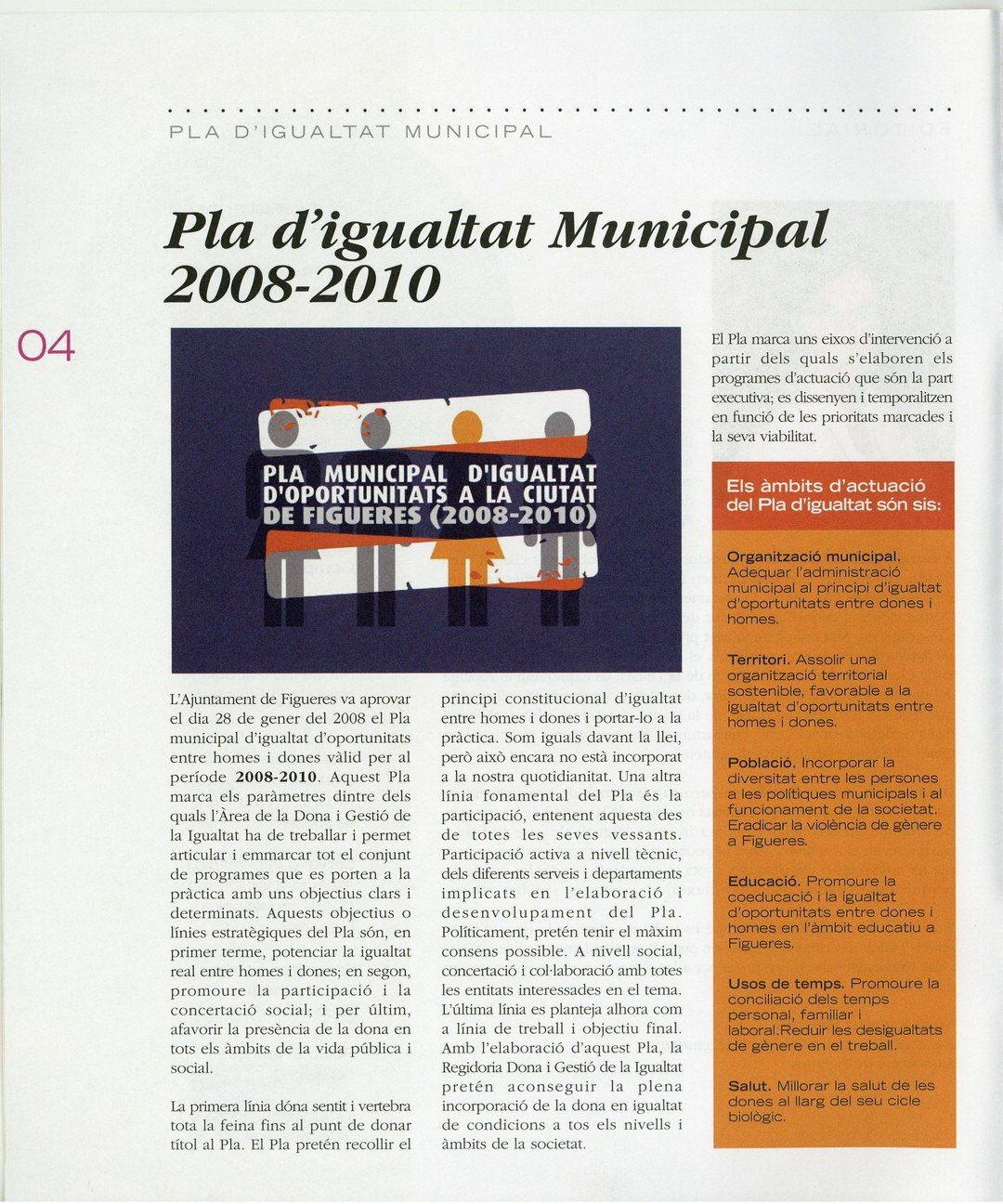}\par 
    \end{multicols}
\caption{Example of different newspapers from the XAC-Newspapers dataset; years 1905 (left) 1982 (middle) and 2009 (right).}
\label{fig:xac_teaser}
\end{figure*}

The second kind of document we have used to evaluate our method has been Newspaper date estimation \ref{fig:xac_teaser}. This data collection was gently yielded by the archivist from the Xarxa d'Arxius Comarcals \footnote{XAC is a governmental archivist institution with url for further detail: \url{https://xac.gencat.cat/en/inici/}}; belonging to the National Archives of Catalonia. This dataset consists of 10,001 newspaper pages from the year 1847 to 2021. Figure \ref{fig:xac_histogram} shows some samples.
In our experiments we gray-scaled all the images to prevent a color bias in the retrieval process.

Due historical and practical reasons the dataset is heavily unbalanced and sparsely distributed. Nevertheless almost every period contains some samples.

\section{Learning Objectives}
    \label{sec:learning_obj}
Following our previous work for word spotting \cite{riba2021learning} and scanned image retrieval~\cite{molina2021date}, our document dating model uses the smooth-nDCG (Equation \ref{eq:DCG}) as its learning objective function. The smooth-nDCG is a differentiable information retrieval metric. Given a query $q$ and the set of retrieved elements from the dataset \(\Omega_q\), the smooth-nDCG is defined as follows:.

\begin{equation}
    \operatorname{DCG}_q \approx \sum_{i\in \Omega_q} \frac{r(i)}{\log_2\left(2 + \sum_{j\in \Omega_q, j \neq i} \mathcal{G}(D_{ij};\tau)\right) }
    \label{eq:DCG}
\end{equation} %

\noindent where $r(i)$ is a graded function on the relevance of the \textit{i}-th item in the returned ranked list, $\mathcal{G}(x;\tau) = {1}/({1+e^\frac{-x}{\tau}})$ is the sigmoid function with temperature $\tau$ pointing out if element $i$ is relevant with respect $j$ as a mimic for binary step function, and $D_{ij} = s_i - s_j$, with $s_i$ and $s_j$ being respectively the cosine similarity between the \textit{i}-th and \textit{j}-th elements to the query.

\begin{figure}
    \centering
    \includegraphics[scale=0.5]{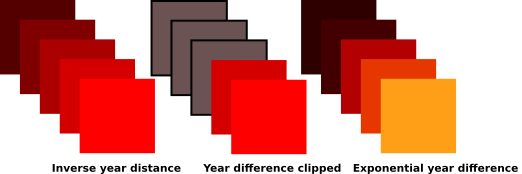}
    \caption{Example of different relevance functions $r(i)$ that could be applied to a set of data.}
    \label{fig:relevances}
\end{figure}

This function measures the quality of retrieved content given a query but the retrieved content may be graded in a decreased scale. For example, in the case of document dating, it does not make sense to use a binary relevance function in which the documents of the same year as the query are considered relevant and the rest are considered not relevant. Instead, we would like to have a graded relevance function ($r(i)$) that ranks elements by their date-distance to the query (i.e. elements 1, 2 or 3 years away might be relevant in this order).
This is reflected in equation~\ref{eq:DCG} by the numerator $r(i)$ meaning the relevance of the document $i$ with respect the query $q$. This can be considered the ground truth label for the problem to optimize. In Figure \ref{fig:relevances} we illustrate graphically different relevance functions that prioritize retrieving images from closer dates (yellow and red indicates higher values for the relevance function $r(i)$ and darker colors indicates lower values).

Note that the relevance $r(i)$ given a query $q$ can be represented by a relevance matrix where $M_qi$ is the relevance of the document $i$ with respect the query $q$. This can be saw as an attention matrix, what may be interesting to observe.

Since the document dates' annotations are $\mathbb{R}^1$ scalars (years), it makes sense to measure the distance between documents with their dates' absolute difference. In this paper we use various slightly different formulations of $r(i)$: $\gamma$-thresholded difference (Equation \ref{eq:relevance}), the logarithmic scaled difference (Equation \ref{eq:relevanceLog}),  or a combination of them. In order to keep the distances in a reasonable range; we used Equation \ref{eq:relevance} for problems with a few categories such as the newspaper retrieval, where there's only a range of one century. When distances may return too big numbers, we scale it with Equation \ref{eq:relevanceLog} such as MPS dataset, which has a wider range $[min_{year}, max_{year}]$.

\begin{align}
    r(n;\gamma) & = \max(0,\gamma - | y_q - y_n |) \label{eq:relevance}\\
    r(n) &= \log{({1+| y_q - y_n|})} \label{eq:relevanceLog}
\end{align}%

\section{Proposed Method}
    \label{sec:arch}
As mentioned above, to the best of our knowledge the existing methods for date estimation an retrieval can not be generalized to different types of document datasets. There's works that better meet this requirement such as He Sheng \textit{et al.} \cite{he2016Multiple} \cite{HE2016159} that requires a certain priory propositions; or Muller's work \cite{muller2017picture} that since it is based in categorical classification it can not predict if a manuscript classified of a certain period belongs to the early or late registers in it. Nevertheless it is notable that some methods such as Hamid's \textit{et. al} \cite{hamid2019deep} have shown an extraordinary performance in terms of strict date estimation without any a priory information. On the other hand, since this method relies on information retrieval; many applications and variations may emerge. This is the case of the application we are presenting.

\begin{figure}
    \centering
    \begin{tabular}{c c}
    \includegraphics[height=0.24\linewidth ]{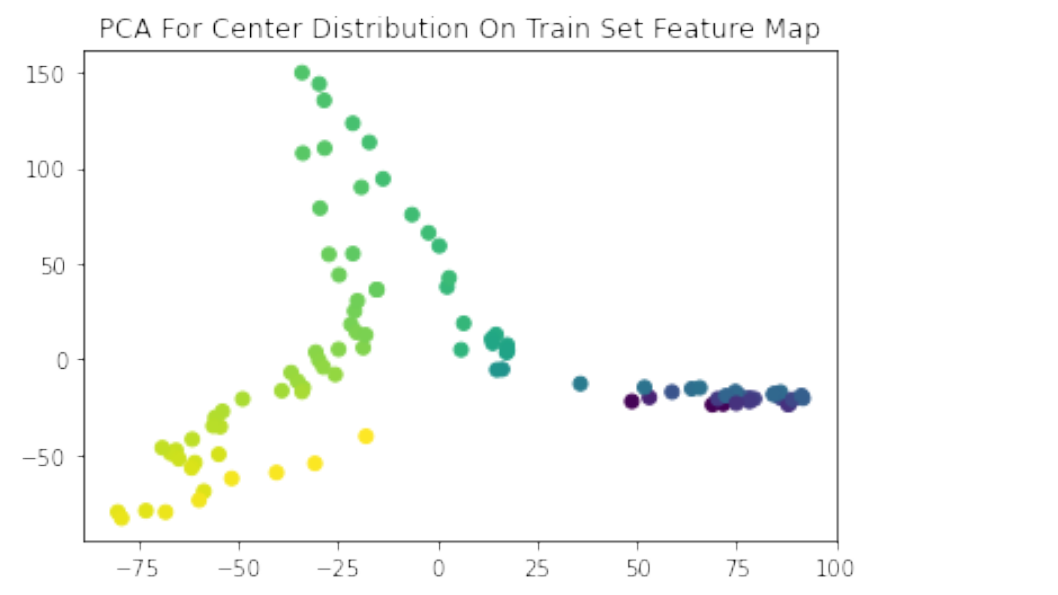}
    \includegraphics[height=0.24\linewidth ]{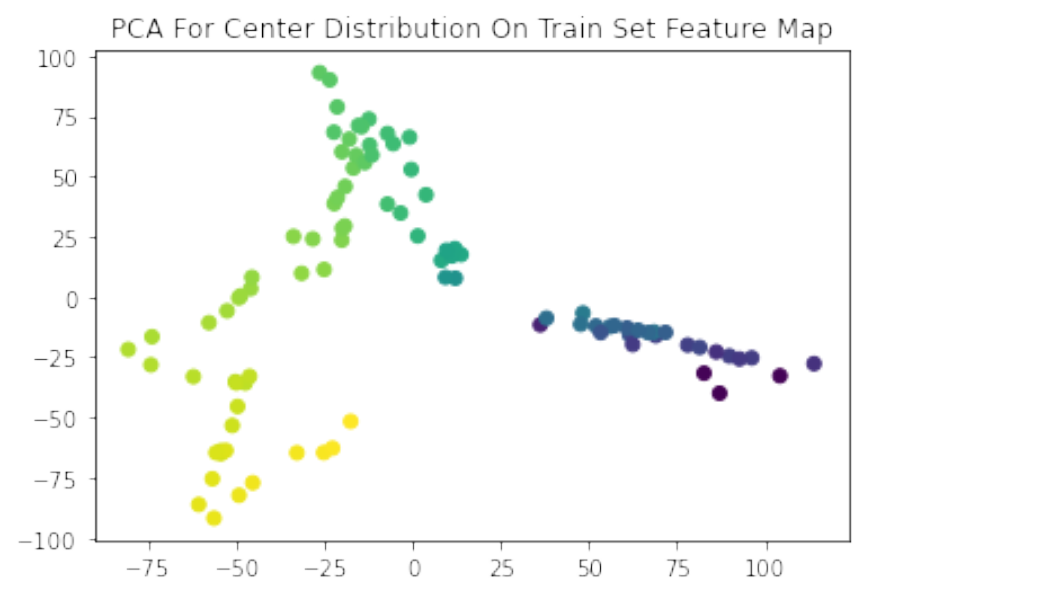}
    \end{tabular}
\caption{Main 2D Embedding space (left) compared to an specialized in early years embedding space (right).}
\label{fig:retrain}
\end{figure}

Using the benefits of being able to effectively retrieve information from different layers of relevance, we can build a Human-in-the-loop system that uses user feedback to adapt the specificity of the use case Figure \ref{fig:retrain}. As it will be discussed in Section \ref{sec:experiments}; the importance of the prediction error decays in favor of adaptability not only to different datasets but to focusing on different labels in the same one.

\begin{figure}
    \centering
\begin{tabular}{c c c}
    \includegraphics[height=0.20\linewidth ]{images/emb1} & 
    \includegraphics[height=0.20\linewidth ]{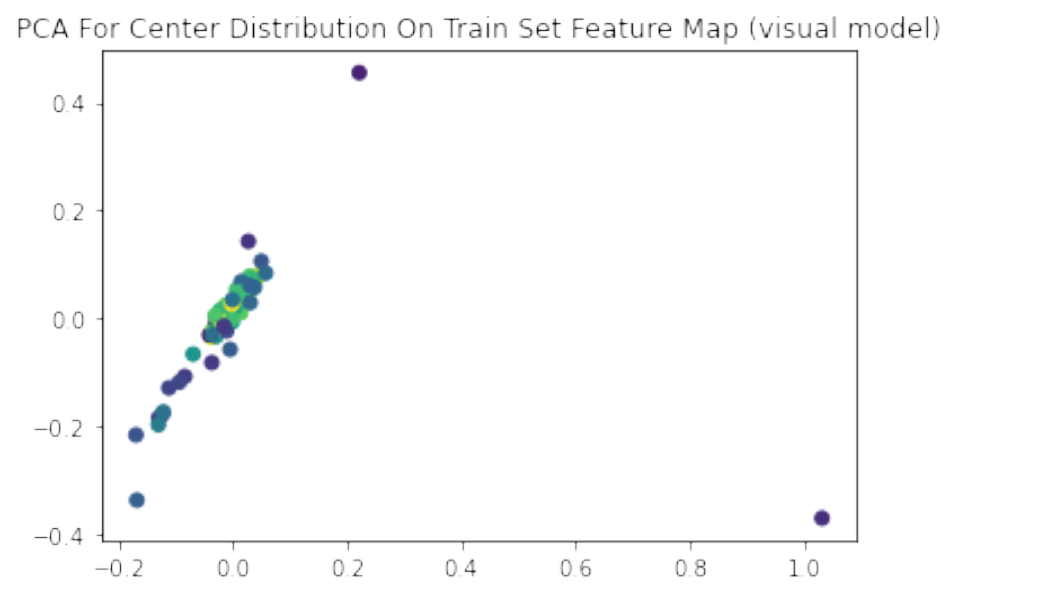} & 
    \includegraphics[height=0.20\linewidth ]{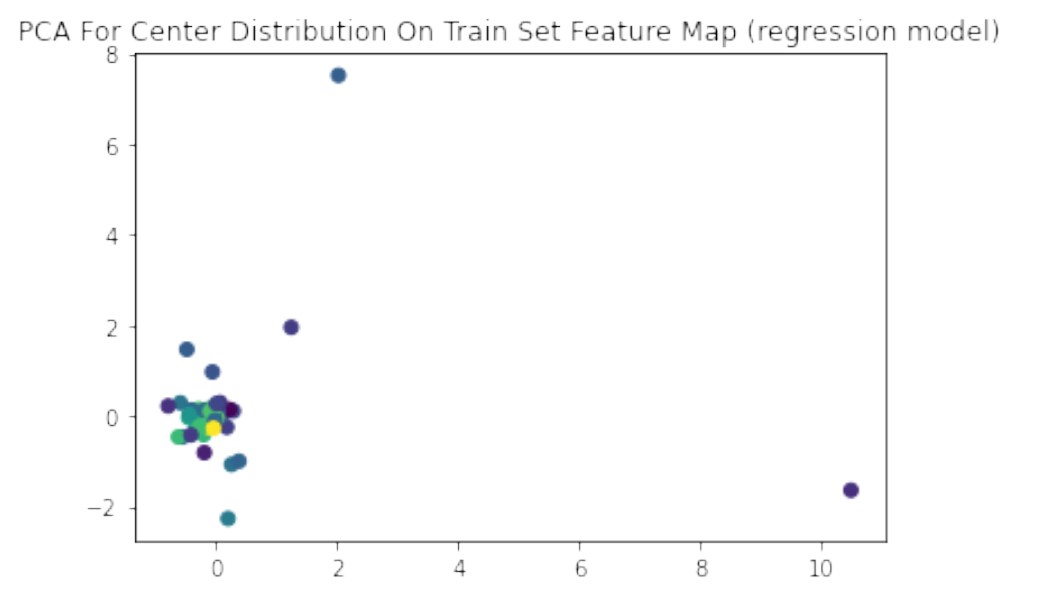}
\end{tabular}
    \caption{PCA 2D Projection for Newspapers Dataset cluster centers on (from left to right): nDCG model, Imagenet weights, regression model from \ref{tab:baselines_xac}. Color gradient is proportional to label gradient (years).}
    \label{fig:emb}
\end{figure}

Given a labeled dataset, we train a Convolutional Neural Network (CNN) to learn a document projection to an embedding space where the distance between points is proportional to their distance in years, not specific visual features or content. This is made by maximizing the smooth-nDCG function; a normalization of Equation \ref{eq:DCG}. Algorithm~\ref{alg:train} describes the training algorithm for the proposed model. 
Once the neural network is able to sort documents according to their actual dates; we can project a bunch of unlabeled ones expecting it to be properly clustered as a sorting Figure \ref{fig:emb}.

\begin{figure}
    \centering
    \includegraphics[scale=0.55]{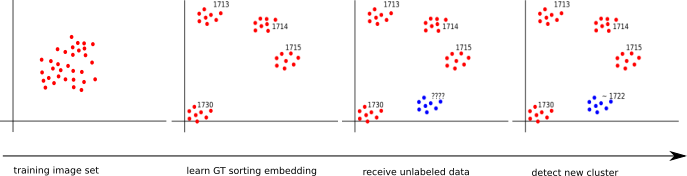}
    \caption{Method pipeline for predicting years outside the training set. Since the clusters are sorted optimizing the ranking function; a new cluster found between two ones could be labeled as the mean between both clusters.} 
    \label{fig:pipeline}
\end{figure}

With this method, if we have enough continuity and density of data so the "years" space is organized in cluster such as $Y = \{ 1700, ..., 1715\} \cup \{1730, ...\}$, the method can figure out that if the new bunch of data is placed between two clusters. Figure \ref{fig:pipeline} illustrates this concept of inter-cluster embedding. We can use cosine distance to perform a weighted prediction for this new cluster. Note that given a new bunch of data many points could already belong to a certain cluster while many others could not, forming then new ones.

So, as it has been presented before, with smooth-nDCG date estimation any prior information is not required, and despite not having certain labels, we can find them out by properly exploring the cluster distribution with respect to the new data.

\begin{algorithm}[H]
\hspace*{\algorithmicindent} \textbf{Input:} Input data $\{\mathcal{X}, \mathcal{Y}\}$; CNN $f$; distance function D; max training iterations $T$ \\ 
\hspace*{\algorithmicindent} \textbf{Output: } Network parameters $w$
\begin{algorithmic}[1]
    \State Initialize relevance matrix $R: |\mathcal{Y}|\times|\mathcal{Y}|$
    \For{$y_q \leftarrow$ year in $\mathcal{Y}$}
        \For{$y_n \leftarrow$ year in $\mathcal{Y}$}
        
            \State Calculate relative relevance $R_{i, j} \leftarrow $  Eq. \ref{eq:relevance}
    
        \EndFor
    \EndFor
    \Repeat
    \State Process images to output embedding $h \leftarrow f_w(\{x_i\}_{i=1}^{N_{batch}})$
    \State Get Distance matrix from embeddings, all vs all rankings $M \leftarrow D(h)$
    \State Get relevance from relevance matrix given a query $q$, $r(i)\leftarrow R_{i, q}$ 
    \State Using the relevance score, $\mathcal{L} \leftarrow$ Eq. ~\ref{eq:DCG}
    \State $w \leftarrow w + \eta(\nabla_{w} \mathcal{L})$ \label{alg:line:H}
    \Until{Max training iterations $T$}
\end{algorithmic}
\caption{Training algorithm for the proposed model.} \label{alg:train}
\end{algorithm}

\section{Application}
    \label{sec:experiments}
\subsection{Smooth-nDCG Human-In-The-Loop Architecture}
\begin{figure}
    \centering    \includegraphics[width=11cm,height=6cm]{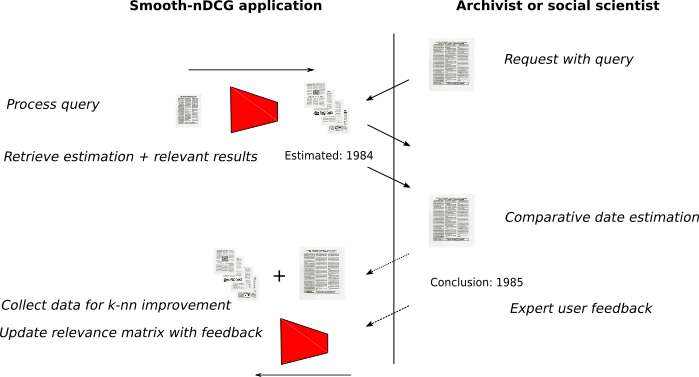}
    \caption{Communication between user and a smooth-nDCG application. Observe that it's not an unilateral communication scheme.}
    \label{fig:coms}
\end{figure}

As commented above we propose an application that is designed as a service to experts that deal with lot of unlabeled data that they have to sort, label, or extract some cues of information such as location or keywords. As illustrated in the application-level scheme in Figure \ref{fig:coms} our solution is generic and applicable to any document domain and dataset, the only module that needs to be changed is the relevance matrix of the data.

\begin{figure}
    \centering
    \setlength{\tabcolsep}{1.2pt}
    \begin{tabular}{ccccc}
        \textsf{\scriptsize Query / 1350} & \textsf{\scriptsize Response 1 / 1375} & \textsf{\scriptsize Response 2 / 1375} & \textsf{\scriptsize Response 3 / 1375} & \textsf{\scriptsize Response 4 1350}\\
        \includegraphics[width=0.15\linewidth]{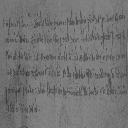} & \includegraphics[width=0.15\linewidth]{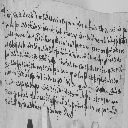} & \includegraphics[width=0.15\linewidth]{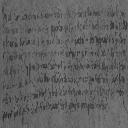} & \includegraphics[width=0.15\linewidth]{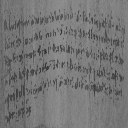} & \includegraphics[width=0.15\linewidth]{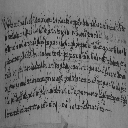}\\
        \textsf{\scriptsize Response 5 / 1375} & \textsf{\scriptsize Response 6 / 1375} & \textsf{\scriptsize Response 7 / 1350} & \textsf{\scriptsize Response 8 / 1375} & \textsf{\scriptsize Response 9 1350}\\
        \includegraphics[width=0.15\linewidth]{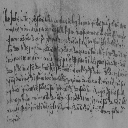} & \includegraphics[width=0.15\linewidth]{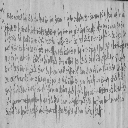} & \includegraphics[width=0.15\linewidth]{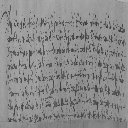} & \includegraphics[width=0.15\linewidth]{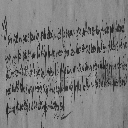} & \includegraphics[width=0.15\linewidth]{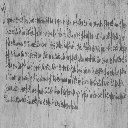}\\
    \end{tabular}
    \caption{Qualitative results for a query from the Medieval Paleographic Scale (MPS) test set \cite{he_sheng_2016_1194357}.}
    \label{fig:CharQualitative}
\end{figure}

We appreciate that our main intention is being able to retrieve useful information for the user, this is where the retrieval function shines brighter than classification approaches. Once retrieved the information with which the user should be able to perform their task (such as comparative date estimation in this case) they can improve their own application instance by joining its newly labeled data to the set for the k-nn selection. As shown in Figure \ref{fig:CharQualitative}
given a query we can suppose unlabeled, the top of retrieved documents should help the user to run a series of conclusions by comparison; since the entire response is labeled, they should not start from scratch, but from a good benchmark given by the k-nn approximation. 

Additionally, as it was exposed in Algorithm \ref{alg:train}, we are using a relevance matrix that tells how relevant is a category, or a certain data to another. In our application this property can  be used to incorporate feedback from the user. For example, the user might not want to consider the distances in the early 20s period as relevant as the distances in the late 90s.  If the user is not interested in studying data from further categories, this should allow the model to train again focusing on the categories the user is currently interested in. This Human-in-the-loop training allows the user to adapt the model to their specific needs in an intuitive manner, without any knowledge about deep learning or computer vision.

\subsection{Quantitaive Evaluation}


We have trained and evaluated our document dating model in the two datasets described in section~\ref{sec:data}. Tables~\ref{tab:baselines_mps} and~\ref{tab:baselines_xac} show the obtained results and a comparison with other existing methods and baselines in terms of Mean Average Error (MAE) and mean Average Precision (mAP). While in the MPS dataset (Table~\ref{tab:baselines_mps}) our model performance is far from the state-of-the-art, we want to emphasise that our primarily goal in this paper is not to obtain a perfect MAE but into design a practical human-in-the-loop application. In that regard it is important to notice that no other model in Table~\ref{tab:baselines_mps} can directly provide a retrieval document set that allows the user to make more informed predictions and at the same time improve the model's performance in an easy way. 
The experiments ran on Tables \ref{tab:baselines_mps}\ref{tab:baselines_xac} were performed with an Inception v3 pre-trained on ImageNet fine-tuning all the parameters. As it will be discussed on Section \ref{sec:conclusions}, further research is needed in terms of optimizing this CNNs. Using patches instead of the whole image as shown in \cite{hamid2019deep} should make it perform better. Many clues in previous work \cite{molina2021date} and Table \ref{tab:baselines_xac} point that smooth-nDCG function should be able to perform with equivalent results to a regression model, with the added value of optimizing the ranking metrics.

We used as well an Inception v3 estimator for baseline in Table \ref{tab:baselines_xac} with same fine-tuning criterion and a $1000 \times 1$ linear layer at the top of the network \footnote{mAP approximated from training bacthes}. We added two frozen parameters: given the network $f: X \rightarrow \mathbb{R}^1$ and the frozen parameters $P = (w, b)$ the computed prediction $h$ is the linear combination $h = (f(x), 1) \cdot P^T$. This parameters could be a learnable layer but for faster convergence we decided to let it frozen at $(10, 2000)$ as Muller \textit{et al.} \cite{muller2017picture} froze the bias as the average of the dataset (1930) in the classifier. We used the whole dataset for training and testing the model.


\begin{table}[h]
\caption{Mean Absolute Error (MAE) comparison of our model with existing methods on the test set of the MPS dataset }
\label{tab:baselines_mps}
\begin{tabularx}{\columnwidth}{Xcc}
\toprule
Baseline & MAE & mAP \\   
\midrule
Fraglet and Hinge Features \cite{he2014towards} & 35.4 & - \\
Hinge Features \cite{bulacu2007text} & 12.20  & - \\
Quill Features \cite{brink2012writer} & 12.10 & - \\
Polar Stroke Descriptor (PSD) \cite{he2016Multiple} & 20.90 & - \\
PSD + Temporal Pattern Codebook \cite{he2016historical} & 7.80 & -\\
Textural Features \cite{hamid2018historical} & 20.13 & -\\
InceptionResnetv2 \cite{hamid2019deep} & 3.01 & - \\
\midrule
Smooth-nDCG Manuscript Retrieval (MPS) & 23.8 & 0.43\\
\bottomrule
\end{tabularx}
\end{table}

\begin{table}[h]
\caption{Results and benchmark for newspaper date estimation dataset}
\label{tab:baselines_xac}
\begin{tabularx}{\columnwidth}{Xcc}
\toprule
Baseline & MAE & mAP \\   
\midrule
Regression Baseline (Inception v3) & 3.5 & 0.24  \\ 
Smooth-nDCG Newspaper Date Estimation (Inception v3) & 2.9 & 0.49\\
\bottomrule
\end{tabularx}
\end{table}

\section{Conclusions}
    \label{sec:conclusions}
As exposed in Sections \ref{sec:Introduction}, \ref{sec:experiments}; the retrieval task is a powerful tool for archivists and social scientists. Despite further research is needed in terms of this particular CNN optimization (as Hamid \textit{et al. }\cite{hamid2019deep} proved, Inception can outperform our current MAE error), the designed model provides lot of powerful interactions with the user. The most important idea of this work, is the fact commented in Section \ref{sec:Introduction}; the date estimation task is a very specific one as many others. This means that knowing the prediction of the year of a document may not be useful (which is indeed useful as cue for this specific task). Alternatively, given a query, retrieving a set of labeled images of the same context than the query, can be more useful in the study of a document collection.

In comparison to previous work reviewed in Section \ref{sec:sota}, the proposed method allows the user to evaluate topics under the dataset further than strict date estimation as it was designed to. In contrast to classic content based retrieval where visual features such as textures, patterns or keywords are used for indexing, our proposed system offers the functionality of indexing by historical context. Note that the context is something that emerges directly from the visual content, but by not biasing the model to look for one (or a combination of them) of the commented visual cues, the model gets an optimized representation of each year in the embedding space. This returns us to the previous statement, the fact that the model is learning a continuous efficient representation of each one of the periods (by using a graded retrieval function as loss) \cite{riba2021learning} \cite{molina2021date}, the answers you can get by analyzing the retrieved content are wider than they seem.

The third key point of the application is the malleability and adaptability of the system to new requests and data. As shown in Section \ref{sec:arch}, the usage of a relevance matrix allows the users to provide feedback that improves the system just by tuning the values in the matrix to their convenience i.e. increasing the values of relevance for categories that fit their interest. In the case of date estimation, if there's an interest for a certain year, it's easy to just increase relevance near $R_{i}$. As illustrated in Figure \ref{fig:coms} the second part of this, actually Human-in-the-loop, architecture relies on the improvement of the k-NN estimator by feeding the database with recently labeled data.

In conclusion, this method should pave the way to may social sciences usages for big historical databases or institutions as national archives as mentioned in Section \ref{sec:Introduction}. Nevertheless, as it was demonstrated \cite{hamid2019deep}, the improvement of the hyper-parameters, usage of patches and, in general, better training is an immediate future work. 

\section*{Acknowledgment}

This work has been partially supported by the Spanish projects RTI2018-095645-B-C21, and FCT-19-15244, and the Catalan projects 2017-SGR-1783, the Culture Department of the Generalitat de Catalunya, and the CERCA Program / Generalitat de Catalunya.



\bibliographystyle{splncs04}
\bibliography{refs}

\end{document}